%% file: main.tex
\documentclass[10pt,twocolumn,letterpaper,dvipsnames]{article}

\usepackage[pagenumbers]{cvpr} %

\input{preamble}

\usepackage{url}
\usepackage{graphicx}
\usepackage{amsmath}
\usepackage{amssymb}
\usepackage{booktabs}
\usepackage{multirow}
\usepackage{caption}
\usepackage{subcaption}
\usepackage{tabularx}
\usepackage{colortbl}
\usepackage{xcolor}
\usepackage{wrapfig}
\usepackage{sidecap}
\newcommand{\ours}{SILC}
\newcommand{\oursp}{SILC*}

\newcommand{\ourclip}{SILC-C}
\newcommand{\ourclipp}{SILC-C*}
\newcommand{\oursig}{SILC-S}
\newcommand{\oursigp}{SILC-S*}

\newcolumntype{Y}{>{\centering\arraybackslash}X}
\newcommand{\myparagraph}[1]{\vspace{2pt}\noindent{\bf #1}}
\newcommand\blfootnote[1]{%
  \begingroup
  \renewcommand\thefootnote{}\footnote{#1}%
  \addtocounter{footnote}{-1}%
  \endgroup
}

\definecolor{cvprblue}{rgb}{0.21,0.49,0.74}
\usepackage[pagebackref,breaklinks,colorlinks,citecolor=cvprblue]{hyperref}

\def\paperID{****} %
\def\confName{CVPR}
\def\confYear{2024}

\title{SILC: Improving Vision Language Pretraining with Self-Distillation}

\author{
    Muhammad Ferjad Naeem\,\textsuperscript{1$\star$} \quad
    Yongqin Xian\,\textsuperscript{2*} \quad
    Xiaohua Zhai\,\textsuperscript{3*} \\
    Lukas Hoyer\,\textsuperscript{1,2$\circ$}\quad
    Luc Van Gool\,\textsuperscript{1}\quad
    Federico Tombari\,\textsuperscript{2,4}\\
   \textsuperscript{1}\,ETH Zurich \enskip
   \textsuperscript{2}\,Google \enskip
   \textsuperscript{3}\,Google Deepmind \enskip
   \textsuperscript{4}\,TU Munich\\
  {\tt\small \{mnaeem,lhoyer,vangool\}@vision.ee.ethz.ch, \{yxian, xzhai,tombari\}@google.com}
}

\begin{document}
\maketitle
\blfootnote{$\star$ Research Consultant with Google, * Equal advising, $\circ$ Intern at Google during the project.}

\begin{abstract}
Image-Text pretraining on web-scale image caption datasets has become the default recipe for open vocabulary classification and retrieval models thanks to the success of CLIP and its variants. Several works have also used CLIP features for dense prediction tasks and have shown the emergence of open-set abilities.
However, the contrastive objective used by these models only focuses on image-text alignment and does not incentivise image feature learning for dense prediction tasks. 
In this work, we introduce SILC, a novel framework for vision language pretraining. SILC improves image-text contrastive learning with the simple addition of local-to-global correspondence learning by self-distillation.
We show that distilling local image features from an exponential moving average~(EMA) teacher model 
significantly improves model performance on dense predictions tasks like detection and segmentation, while also providing improvements on image-level tasks such as classification and retrieval.
\ours{} models sets a new state of the art for zero-shot classification, few shot classification, image and text retrieval, zero-shot segmentation, and open vocabulary segmentation. We  further show that SILC features greatly benefit open vocabulary detection, captioning and visual question answering.
\end{abstract}

\begin{figure}
    \centering
    \includegraphics[width=\linewidth]{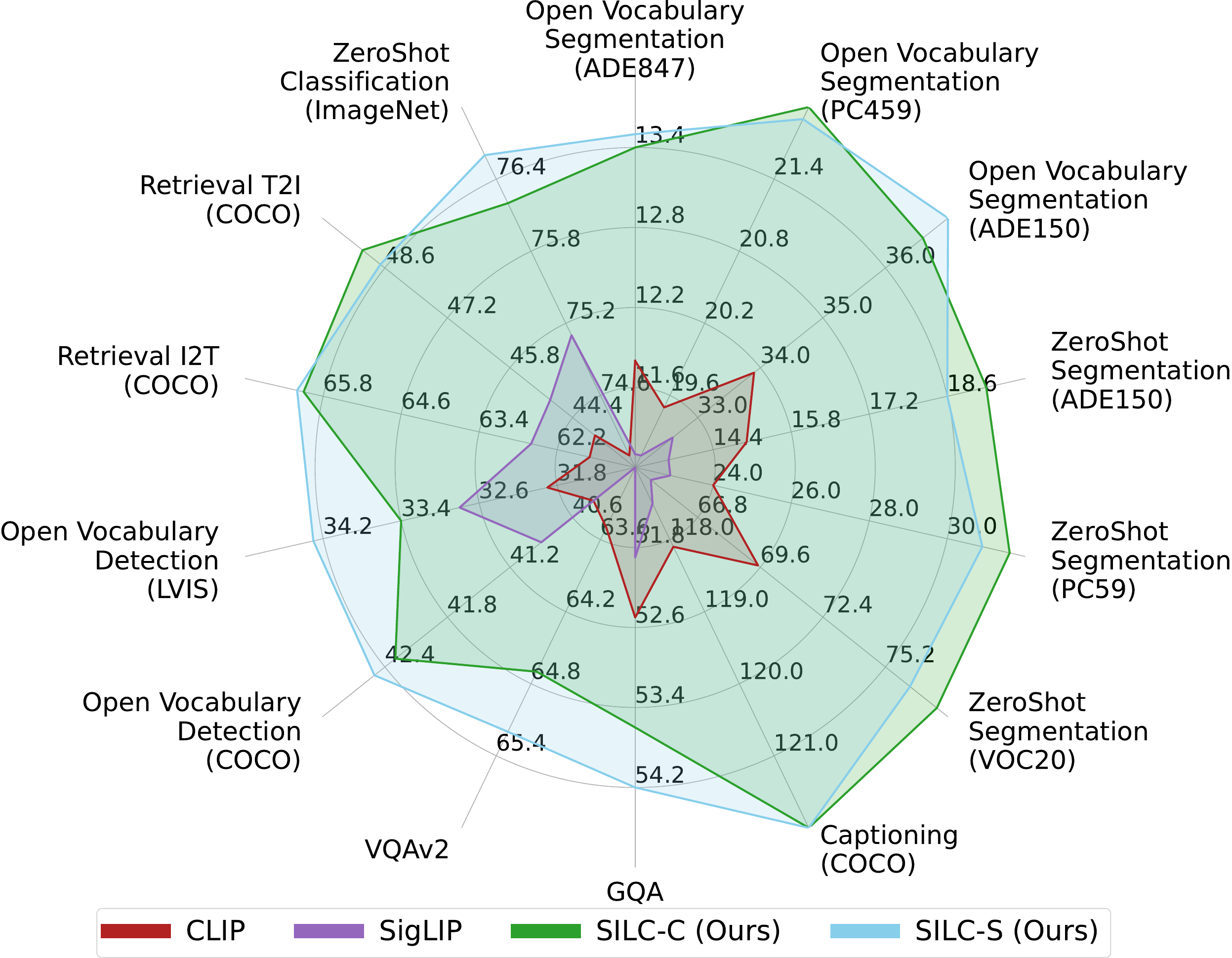}
    \caption{SILC improves image-text contrastive learning with the addition of local-to-global correspondence learning by self-distillation. As a result, SILC models learn more locally aware visual features that are also grounded in language. SILC models offer significant improvements over CLIP~(WebLI) and SigLIP over a wide variety of computer vision tasks including classification, segmentation, detection, captioning, VQA and retrieval.}
    \label{fig:radar}
\end{figure}

\section{Introduction.}
Recent advancements in self-supervised learning~\citep{dino, dinov2, simclr, byol} and weakly supervised learning on web data~\citep{clip, align, zhai2023sigmoid} has spearheaded the development of foundational language~\citep{gpt, chowdhery2022palm} and vision-language models~\citep{clip, align, zhai2023sigmoid}. These methods get around the long term challenge of obtaining large labelled dataset by developing self-supervision objectives. 
Developing open vocabulary computer vision models that can reason beyond a pre-determined set of classes has been a long-term challenge. The introduction of web image-text datasets and the progress in compute have enabled significant advances in this field. Popularized by CLIP~\citep{clip}, contrastive pretraining utilizes large datasets with paired image and text from the web and trains a vision-language model~(VLM) to embed them to a shared latent space. Since these models are trained on a wide set of concepts, the learned VLM allows for open vocabulary inference~\citep{clip}.
However, developing open vocabulary dense prediction models for segmentation and detection is still an open challenge,
since internet-scale datasets do not have dense pixel-level labels.
Several works have found that incorporating VLMs in segmentation and detection models can unlock some open vocabulary abilities~\citep{cho2023cat, ding2022decoupling, xu2022simple, fvlm, convsdie}. Since CLIP is not trained for these tasks, these methods get around its limitations by tuning the learned model with some dense prediction labelled dataset. One set of methods utilizes a normal segmentation / detection model for class agnostic inference and then predict the class logits with CLIP~\citep{cho2023cat, liang2023open}. Another family of methods aims to  distill VLMs directly into a dense prediction model and utilize the text transformer to generate the class weights to predict logits~\cite{lseg, ghiasi2022scaling}. These works have been highly impactful towards expanding open vocabulary abilities of dense prediction models. However, since the contrastive pretraining objective does not explicitly encourage learning good local features for dense prediction tasks, these methods are limited by the VLM's intrinsic performance~\citep{dinov2} as we also show later in our experiments. 

In the self-supervised literature, enforcing local-to-global consistency by self-distillation has emerged as a powerful pretraining objective~\citep{dino, dinov2, ibot} to learn vision backbones that are competitive on classification as well as dense prediction tasks, e.g. segmentation and detection. However, these backbones can not directly be used for zero-shot or open vocabulary inference as they do not contain any notion of class or language in the model.
 In this work, we propose SILC, which combines the advantages of these two branches and unifies image-text contrastive pretraining and local-to-global consistency learning.
\ours{} utilises a web image-text dataset to learn one model that improves VLM performance on existing classification and retrieval tasks while especially improving performance on zero-shot and open vocabulary segmentation, open vocabulary detection, captioning and Visual Question Answering~(VQA).

Our contributions are as follows:
1. We propose a novel training framework for VLMs that pairs contrastive pretraining on image-text data with self-distillation on web images.
\new{2. While conceptually very simple, we show that by learning stronger visual features with better local understanding, \ours{} models offer consistent improvements on multitude of computer vision tasks. These improvements are especially apparent on tasks that require better local understanding including zero-shot segmentation, open vocabulary segmentation, open vocabulary detection, captioning and Visual Question Answering~(VQA).}
3. We contribute a new foundation model that sets a new state of the art on zero-shot classification, few-shot classification, image-to-text and text-to-image retrieval, zero-shot semantic segmentation and open vocabulary semantic segmentation.

\section{Related Works.}

\myparagraph{Image-Text Pretraining.}
Vision-language model~(VLM) pretraining~\citep{clip, align, blip, pali} aims to learn generic multimodal representations that generalize to a wide range of downstream tasks. Substantial progress has been made in this field towards better pretraining objectives~\citep{align, simvlm} and better large-scale image-text dataset~\citep{clip, pali}. One of the most popular objective functions is contrastive learning~\citep{clip, align} that pulls positive image and text pairs close and pushes negative ones apart in the joint embedding space. It is capable of scaling to a large-scale pretraining dataset and learning highly discriminative image and text features. Many works~\citep{declip, lit, zhai2023sigmoid, filip, naeem2022i2dformer, naeem2023i2mvformer, evaclip, datafiltering} in this direction have demonstrated improvements across zero-shot image classification and retrieval benchmarks. 

Another line of research focuses on generative learning via autoregressive text generation~\citep{simvlm, git, cappa}. 
Compared to the contrastive learning, generative learning usually performs better on text generation tasks e.g., image captioning and VQA. Finally, there are  hybrid methods~\citep{flamingo, albef, flava, vilbert, coca, blip} that combine multiple objective functions including generative, contrastive and multi-task losses. While many VLMs~\citep{clip, simvlm} mainly focus on learning global image-text alignment that benefit image-level downstream tasks, our work aims to develop a new VLM that benefits both image-level and pixel-level tasks. There have been a few attempts~\citep{fiber, segclip, regionclip, dong2023maskclip} to improve VLMs for dense prediction tasks including object detection and semantic segmentation. However, they are either modeling the fine-grained patch-text interactions that are not scalable~\citep{fiber, segclip} or rely on additional bounding box annotations~\citep{regionclip, glip}. 
In this work we propose to pair image-text contrastive learning with self-distillation to learn a VLM.

\myparagraph{Self-supervised Learning.}
Self-supervised learning is another popular pretraining paradigm where features are learned from image data itself.
 One branch of methods optimize the network to solve pretext tasks e.g., image coloring~\citep{colorful}, inpainting~\citep{pathak2016context},  transformation prediction~\citep{gidaris2018unsupervised}, and patch ordering~\citep{misra2020self}. Another family of approaches adopt instance-level discriminative learning via contrastive learning~\citep{simclr, he2020momentum} and clustering~\citep{caron2018deep, caron2020unsupervised}. Recently, \citep{mae} shows that masked autoencoder is also a scalable self-supervised learner. Our work is inspired by DINO~\citep{dino} which shows that segmentation emerges from learning local and global-views consistency. However, DINO cannot be directly used for zero-shot and open-vocabulary inference because it only learns image features. \new{In contrast, our method is trained on image and text data jointly. We show that together with text data, the DINO objective allows the model to develop an understanding of local features and their semantic classes. Therefore our model can potentially directly benefit far more computer-vision applications.}

\myparagraph{Zero-shot Semantic Segmentation.}
Zero-shot semantic segmentation aims to segment arbitrary visual concepts in the wild without dense annotations~\citep{xu2022groupvit}. 
Methods in this area rely on image-text pairs from a combination of image captioning and web image-text dataset. Since these datasets do not have dense labels, they utilize a self-supervised image region to text attention criterion.
Group-VIT~\citep{xu2022groupvit} proposes to introduce grouping tokens that cluster similar image patches under each group token. 
MaskCLIP~\citep{maskclip} and CLIPpy~\cite{clippy} found that normal CLIP training results in zero-shot segmentation emerging. 
ReCo~\citep{shin2022reco} proposes a refinement process on top of MaskCLIP by retrieval and co-segmentation. Finally, the current state-of-the-art TCL~\citep{tcl} learns a decoder to upsample the grounded patch embeddings and learns a region to text attention.

\myparagraph{Open Vocabulary Segmentation and Detection.}
Open-vocabulary semantic segmentation methods aim to segment images according to a vocabulary of class categories provided at test-time containing additional unseen classes. 
In contrast to zero-shot segmentation, open-vocabulary semantic segmentation has access to a semantic segmentation dataset with a limited vocabulary for training. 
Recent methods transfer the open-vocabulary capabilities of CLIP from image- to pixel-level predictions. 
LSeg~\citep{li2022languagedriven} learns pixel-wise visual embeddings that align with CLIP text embeddings while OpenSeg~\citep{ghiasi2022scaling} learns class-agnostic segmentation proposals to pool visual features for region-text grounding. ZegFormer~\citep{ding2022decoupling} and ZSseg~\citep{xu2022simple} introduce a two-stage framework, which first learns class-agnostic segmentation mask predictions and classifies the corresponding region using a frozen CLIP. OVSeg~\citep{liang2023open} further finetunes CLIP on region-text pairs to compensate for the appearance shift of masked crops. To avoid the overhead of two stages, CAT-Seg~\citep{cho2023cat} learns the aggregation of cost volumes between text embeddings and dense image embeddings from CLIP. \new{Towards open vocabulary detection one family of methods, e.g. OWVLv2~\cite{owlvit}, RegionCLIP~\citep{regionclip}, Detic~\cite{detic}, 3Ways~\cite{threeways}, pseudolabel boxes for image caption data to use for localization pretraining. An orthogonal family of methods including~\cite{glip, glipv2, detclip, detclipv2} pretrain models to align class agnostic pseudoboxes to text as pretraining.}

\begin{figure}
    \centering
    \includegraphics[width=\linewidth]{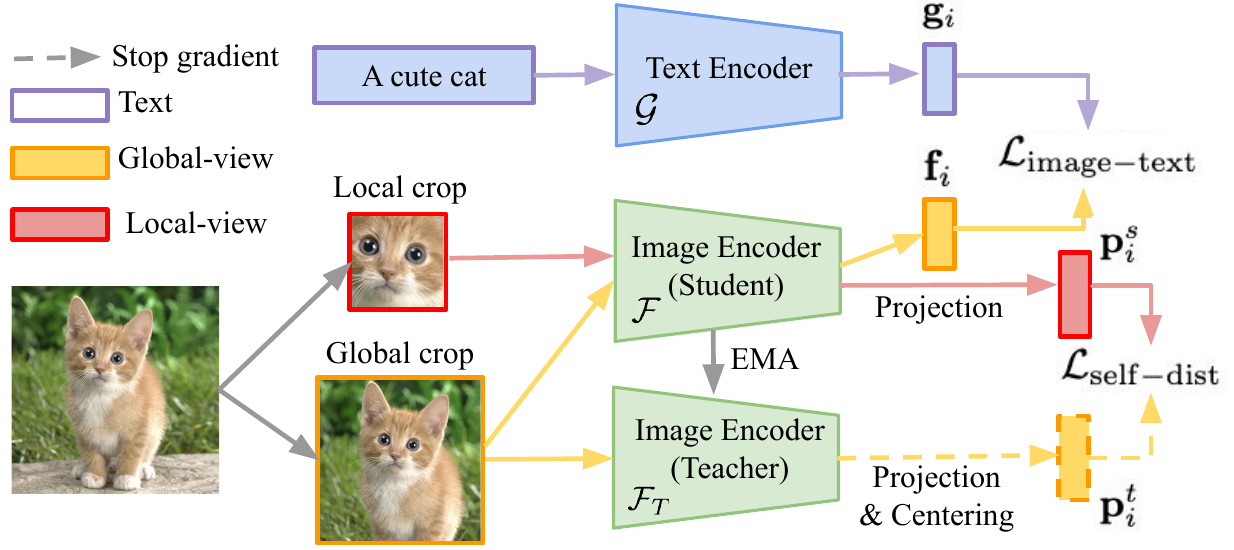}
    \caption{\textbf{\ours{}} is a two-tower transformer based VLM. The first component of our training objective uses a global view of an image covering a large area and its paired caption to optimise a batch-wise contrastive loss for images and texts. The second component of our training objective enforces local-to-global consistency by self-distillation between the main model (the student) and an Exponential Moving Average~(EMA)-based teacher.  
    This local-to-global correspondence additionally allows the model to learn good visual features. Together the two objectives allow the model to excel at both traditional VLM tasks as well as tasks that require local understanding like segmentation and detection.}
    \label{fig:model}
\end{figure}

\section{Method.}
\ours{} builds on the contrastive pretraining framework of CLIP~\citep{clip} and SigLIP~\citep{zhai2023sigmoid}. \ours{} consists of a two-tower transformer model with a shared embedding space. We utilize a web-scale paired image-text dataset and rely on large-scale pretraining to learn the weights of the model. The first component of our pretraining objective focuses on aligning matching image-text pairs close together and away from other images and texts in the batch. This objective has been incredibly successful in recent literature~\citep{clip, zhai2023sigmoid}. However, the contrastive objective in its current form does not focus on capturing rich local image semantics necessary for dense prediction tasks like segmentation and detection. Therefore, we propose to pair the contrastive pretraining objective with a local-to-global consistency objective that uses self-distillation as shown in Figure~\ref{fig:model}. 
\textbf{\ours{}} gets its name from the two training objectives consisting of \textbf{S}elf-Distillation from Images and \textbf{I}mage-\textbf{L}anguage \textbf{C}ontrastive Alignment from Image-Text pairs.

\subsection{Aligning Image and Text.}
\label{sec:contrastive}
The contrastive pretraining objective relies on the Info-NCE framework~\citep{infonce}. It utilizes large amount of web-scale image-text dataset to learn an alignment between paired image and text. Given a minibatch $\mathcal{B}=\{(I_1, T_1), (I_2, T_2), \dots\}$, where $(I_i, T_i)$ denotes a matching pair of image and text, the contrastive objective encourages matching image and text pairs to lie close together in a shared embedding space. The image $I_i$ is processed by a learnable Vision Transformer $\mathcal{F}$ to get its feature embedding. Similarly, the tokenized text $T_i$ is processed by a learnable Text Transformer $\mathcal{G}$ to get its feature embedding. These feature embeddings are normalized by their $l_2$ norm to get $\mathbf{f}_i=\frac {\mathcal{F}(I_i)} {\|\mathcal{F}(I_i)\|_2} \in \mathbb{R}^{J}$ for the image $I_i$ and $\mathbf{g}_i=\frac {\mathcal{G}(T_i)} {\|\mathcal{G}(T_i)\|_2} \in \mathbb{R}^{J}$ for the paired text $T_i$ where $J$ is the feature dimension of the shared embedding space. \new{The dot product of $\mathbf{f}_i$ and $\mathbf{g}_i$ computes their cosine similarity and is optimized with a pair of cross-entropy losses as proposed by CLIP~\citep{clip} or a sigmoid loss as proposed by SigLIP~\citep{zhai2023sigmoid}. 
The batch-wise contrastive losses of CLIP/ SigLIP, represented as $\mathcal{L}_{\mathrm{image-text}}$, rely on a large batch size to align image-text pairs. This objective tuned over a large amount of data learns a shared embedding space between image and text and thus can be used for zero-shot transfer to multitude of computer vision tasks.}

\subsection{Distilling Local Image Features.}
The image-text contrastive loss has shown to be very successful in learning zero-shot transfer models~\citep{clip, align}. Models learned with this objective have also been used to improve dense prediction tasks like open vocabulary segmentation and detection. However, the contrastive objective alone does not explicitly focus on learning good visual features for dense prediction tasks. These tasks require local image semantics to be sufficiently encoded in the output image and patch embeddings. 
Enforcing local-to-global consistency has emerged as a powerful technique to accomplish this on large unlabelled image data~\citep{dino, dinov2, ibot} in self-supervision literature. \new{However, these methods can not be directly used for open vocabulary models as they are trained without any language information.}
In the second component of our training framework, we take inspiration from this subset of literature and additionally add local-to-global consistency as a training objective for images in our image-text dataset.

The basic idea of this objective is as follows. A teacher network gets a global view of the image representing the scene as a whole and produces a feature embedding. A student model gets a partial view of the same image and produces a feature embedding. A self-distillation objective is introduced where the student needs to match the prediction of the teacher while only having partial information. This enforces the model to learn local semantics and their relation to global semantics of the scene. We add this criterion for the image encoder $\mathcal{F}$. We add a projection as a learnable MLP on top of the image encoder to map from the original shared embedding space of dimension $J$ to $K$ where $K>J$. The student $\mathcal{F}_S$ is the main image encoder with a learnable projection head. Since we rely on noisy web scale image-text data, we do not have an oracle teacher for the student to match. We therefore construct our teacher $\mathcal{F}_T$ as a exponential moving average of the student $\mathcal{F}_S$ from the previous training iterations to realize our self-distillation framework: 
\begin{equation}
 \mathcal{F}_T \leftarrow \lambda \mathcal{F}_T + (1-\lambda) \mathcal{F}_S,
    \label{eq:ema}
\end{equation}
where $\lambda$ controls the update step of the teacher.
For a given image $I_i$, the teacher processes its global crop to produce $\mathbf{p}_i^t \in \mathbb{R}^K$ and the student processes its local crop to produce $\mathbf{p}_i^s\in \mathbb{R}^K$.
To prevent the teacher from collapsing to a trivial solution, we apply sharpening on the outputs of teacher with $\tau_t$ and student with $\tau_s$. To encourage each feature dimension to contribute to the output feature, we additionally introduce a centering operation on the prediction of the teacher. The centering term $\mathbf{c} \in \mathbb{R}^K$ is initialized with $0$ and is updated by a momentum update with a factor of $m$ with the first order batch statistics of the teacher's prediction at each step as follows: 
$\mathbf{c} \leftarrow m \mathbf{c} + (1-m) \frac{1}{|\mathcal{B}|} \sum_{i=1}^{|\mathcal{B}|} \mathbf{p}_i^t$.

To learn local-to-global correspondences, the student is faced with an information asymmetry. The student is given a local view of an image which is realized as a random crop over a small region of the image. The teacher, however, has access to a global view of the image containing more information about the scene. The student is tasked with matching the semantics of the teacher while only having partial information. Therefore, for a given image, the model needs to learn local semantics of the image and how it would fit in the global context of this image. This is realized as a knowledge-distillation loss where the student and the teacher's feature vectors are first converted to a probability distribution by applying a softmax on the teacher prediction $\mathcal{P}_t(I_i^{gl}) = \texttt{softmax}((\mathbf{p}_i^t-\mathbf{c})/ \tau_t)$ and student prediction $\mathcal{P}_s(I_i^{lc}) = \texttt{softmax}(\mathbf{p}_i^s / \tau_s)$.
The student is optimized to match the teacher with a cross-entropy loss,
\begin{equation}
    \mathcal{L}_{\mathrm{self-dist}} = - \mathcal{P}_t(I_i^{gl})^{\intercal}\mathrm{log}(\mathcal{P}_s(I_i^{lc})).
\end{equation}
This self-distillation objective incentivises the image encoder to learn local semantics of images over the large web scale dataset. Since the teacher is constructed with the student's weights, and the image level features are pooled from patch embeddings in a Vision Transformer, this allows for richer local semantics to be captured in the image level as well as the patch level features. 

\new{While this objective has been explored in self-supervised learning~\citep{dino, dinov2}, to the best of our knowledge, we are the first work to show its complimentary nature to image-text contrastive learning on web-scale dataset. We show that when combined with text, this objective allows the model to develop a local understanding of the semantics of an image grounded in language. We find two important modifications compared to previous works that allows it to be complimentary to image-text contrastive learning. 1. Each global view used in $\mathcal{L}_{\mathrm{self-dist}}$ needs to be aligned with text, otherwise the two objectives diverge. This is realized by computing the image-text contrastive loss for each global view while maintaining the same batch size.
2. The momentum scheduler of the EMA should not converge to 1.0. Otherwise the teacher stops learning from image-text loss as the update step becomes too small in the later stage of the training. We therefore use a fixed momentum.}

\setlength{\tabcolsep}{4pt}
\renewcommand{\arraystretch}{1.2} 
\begin{table*}[t]
\centering
\small
\setlength{\aboverulesep}{0pt}\setlength{\belowrulesep}{0pt}
 \resizebox{0.9\linewidth}{!}{%
   \begin{tabular}{ l  c c c c c c  c c c c }
  	\toprule
  	& \multicolumn{2}{c}{\textbf{Zero-Shot Classification}} & \multicolumn{6}{c}{\textbf{Few-shot classification}} & \multicolumn{2}{c}{\textbf{Retrieval}}  \\
  	\cmidrule(lr){2-3} \cmidrule(lr){4-9} \cmidrule(lr){10-11}
   
  \textbf{Model}	
  & \textbf{ImageNet} & \textbf{CIFAR100} & \multicolumn{3}{c}{\textbf{ImageNet}} & \multicolumn{3}{c}{\textbf{CIFAR100}} & \multicolumn{2}{c}{\textbf{COCO}} \\
  \cmidrule(lr){2-2} \cmidrule(lr){3-3} \cmidrule(lr){4-6} \cmidrule(lr){7-9} \cmidrule(lr){10-11} 
  & {T1} & {T1} & {1shot} & {5shot} & {10shot} & {1shot} & {5shot} & {10shot} & {I2T@1} & {T2I@1} \\
        
        \midrule
        \texttt{CLIP (WebLI)}~\citep{zhai2023sigmoid} 
        &  74.1 & 68.4 & 42.8 & 63.2 & 67.3 & 39.4 & 59.6 & 64.6 & 61.7 & 43.9 \\ %
        \textbf{\ourclipp{} ~(Ours)} 
        & \underline{75.3} & \underline{71.0} & \underline{44.6} & \underline{64.3} & \underline{67.8} & \underline{42.8} & \underline{64.6} & \underline{69.6} & \underline{62.5} & \underline{44.9} \\ %
        \textbf{\ourclip~(Ours)} 
        & \textbf{76.2} & \textbf{72.3} & \textbf{45.3} & \textbf{65.0} & \textbf{68.5} & \textbf{45.2} & \textbf{66.9} & \textbf{71.3} & \textbf{66.1} & \textbf{49.1} \\ %
        \midrule
        \texttt{SigLIP}~\citep{zhai2023sigmoid} 
        & 75.1 & \underline{69.8} & 44.0 & 64.2 & {68.4} & 39.0 & 61.7 & 66.3 & 
        {62.6} & \underline{44.9} \\ %
        \textbf{\oursigp{}{(Ours)}} & \underline{75.8} & 69.2 & \underline{45.2} & \underline{64.6} & \underline{68.4} & \underline{40.3} & \underline{63.3} & \underline{67.4} & \underline{63.0} & 44.6 \\ 
        \textbf{\oursig{}(Ours)} & \textbf{76.6} & \textbf{70.6} & \textbf{45.9} &  \textbf{65.2} & \textbf{68.9} & \textbf{41.8} & \textbf{64.9} & \textbf{68.9} & \textbf{66.2} & \textbf{48.7} \\
  	\bottomrule
  \end{tabular}
}
\caption{
\new{\textbf{Comparing \oursp{} with baselines}, we observe that our pretraining framework results in a significant improvement over both CLIP and SigLIP objectives. We reproduce both CLIP and SigLIP on the same WebLI dataset~\citep{pali} to quantify the improvements from our proposed training objective. We further finetune \oursp{} on a cleaner subset to get our final model \ours{} and see that it unlocks additional performance without significant extra retraining. The best performance for each variant is \textbf{bolded}, the second best is \underline{underlined}.  }
}
\vspace{-10pt}
\label{tab:baseline}
\end{table*}

\section{Experiments.}
We compare our \ours{} pretraining framework with both CLIP~\cite{clip} and SigLIP~\cite{zhai2023sigmoid} on the same test bench and perform extensive experimentation. \new{\ours{} models based on the CLIP objective are represented by \textbf{\ourclip{}} and the SigLIP versions are represented by \textbf{\oursig{}}}. We show that \ours{} sets a new state of the art on a variety of tasks: zero-shot classification, few-shot classification, retrieval, zero-shot segmentation and open vocabulary segmentation. \new{We further show that \ours{} models also improve other local semantic understanding tasks including open vocabulary detection, captioning and VQA.}
\subsection{Implementation Details.}
We implement our model in jax in the {\tt big\_vision} codebase~\citep{big_vision,big_vision2}, following the contrastive pretraining setups from~\citep{zhai2023sigmoid}, and use the WebLI dataset\citep{pali} for our experiments. We utilize two global views cropped between $(0.4-1.0)$ of the original image area and eight local views cropped between $(0.05-0.4)$ of the original image area for the self-distillation loss. The global views are resized to $(256\times256$) and the local views are resized to $(96\times96$). 
The teacher momentum $\lambda$ is kept fixed at $0.966$ and the center update momentum $m$ is kept fixed at $0.9$ through the training. The teacher temperature $\tau_t$ is fixed at 0.04 and the student temperature $\tau_s$ is fixed at $0.1$. $K$ is $65536$.
We resize the original image to $(256\times256$) for the contrastive loss between image-text pairs. 
We trained with a batch size of 16k on Google TPUs.
We use \textit{example-seen} to represent how many image and text pairs are drawn from the dataset throughout the training. We train all baselines in our main comparisons in Table~\ref{tab:baseline} for 20 Billion example-seen on the WebLI dataset~\citep{pali} following ~\cite{zhai2023sigmoid}. 
Our models trained on WebLI are marked as \textbf{\oursp{}}.
We use a rsqrt learning scheduler~\cite{scalingvit} with base learning rate of $0.001$ with $50000$ warm up and $50000$ cooldown steps. \new{Additional training details including compute cost are discussed in the supplementary.}

We additionally finetune our model using a smaller but cleaner WebLI subset~\citep{pali} for 1 Billion additional example-seen and represent this model as \textbf{\ours}.
The smaller WebLI subset contains 100 million image-text pairs with finer-grained text filters etc.

\subsection{Classification and Retrieval.}
\new{We compare our pretraining framework with CLIP and SigLIP under the same training and evaluation protocol in Table~\ref{tab:baseline}. We compare at ViT/B16 and see that the introduction of self-distillation to both consistently improve their performance on zero-shot classification, few shot classification and retrieval. On zero-shot classification on ImageNet, \ourclipp{} improves on CLIP (WebLI) by 1.2 points, similarly we notice an improvement of 2.6 points on CIFAR-100 showing the benefit of local feature self-distillation. Similar improvements are noted for few-shot classification where \ourclipp{} improves over CLIP (WebLI) by 1.8, 1.1 and 0.5 points on ImageNet 1 shot, 5 shot and 10 shot classification respectively. We make similar observation on retrieval where \ourclipp{} shows improvements on image to text as well as text to image retrieval. Moving to SigLIP versions of the model, we see a similar trend where the introduction of self-distillation objective allows \oursigp{} to consistently improve almost all metrics over the evaluated tasks. We therefore conclude that capturing better local semantics results in learning stronger visual features which also helps tasks that require global understanding of the image.

Comparing \oursp{} models with \ours{}, we notice that the finetuning on the cleaner subset unlocks additional performance for the model without significant extra training.
For the CLIP based \ourclip{}, We notice another 0.9 point improvement over \ourclipp{} on zero-shot ImageNet classification. We observe improvements of the same magnitude on few-shot classification. Comparing retrieval performance, we see a significant increase in retrieval performance on COCO where \ourclip{} achieves a 3.6 and 4.2 points improvement on Image to Text and Text to Image Recall@1. The SigLIP based \oursig{} follows a similar trend and consistently improves on \oursigp{} on all metrics.  \ours{} models set a new state-of-the-art for these tasks at ViT/B16 model size. 
We also compare with open-source CLIP variants in the supplementary and show \ours{}'s superior performance.

}

\input{qualitatives/zsseg}

\setlength{\tabcolsep}{4pt}
\renewcommand{\arraystretch}{1.2} 
\begin{table}[t]
\centering
\small
\setlength{\aboverulesep}{0pt}\setlength{\belowrulesep}{0pt}
\resizebox{\linewidth}{!}{%
   \begin{tabular}{ l c c c c c c}
  	\toprule
  	\textbf{Model} & \textbf{A-150}  & \textbf{PC-59} & \textbf{Cityscapes} & \textbf{VOC-20} & \textbf{COCO-Stuff} \\
  	\midrule
        \texttt{GroupVIT}~\citep{xu2022groupvit} & 9.2 & 23.4 & 11.1 & \textbf{79.7} & 11.1 \\
        \texttt{MaskCLIP}~\citep{maskclip} & 9.8 & 26.4 & 12.6 & 74.9 & 16.4 \\
        \texttt{ReCo}~\citep{shin2022reco} & 11.2 & 22.3 & 21.1 & 57.7 & 14.8 \\
        \texttt{TCL}~\citep{tcl} & 14.9 & 30.3 & 23.1 & 77.5 & 19.6 \\
        \midrule
        \texttt{CLIP (WebLI)}~\citep{zhai2023sigmoid} & 15.0 & 24.0 & 22.6 & 69.5 & 15.0 \\
        \textbf{\ourclipp{}~(Ours)} & 17.2 & 29.3 & 25.1 & 73.5 & 18.2 \\
        \textbf{\ourclip{}~(Ours)} & \textbf{19.3} & \textbf{31.6} & \textbf{26.9} & 77.5 & \textbf{20.8} \\
        \midrule
        \texttt{SigLIP}~\citep{zhai2023sigmoid} & 13.6 & 22.9 & 20.8 & 64.7 & 13.4 \\
        \textbf{SILC-S*~(Ours)} & 16.7 & 28.6 & 23.4 & 72.1 & 17.3 \\
        \textbf{SILC-S~(Ours)} & 18.6 & 30.9 & 25.2 & 76.3 & 19.7 \\
  	\bottomrule
  \end{tabular}
}
\vspace{-5pt}
\caption{\textbf{Comparing Zero-Shot Segmentation performance} we see that \oursp{} models trained on noisy web image-text data already outperform several ZS segmentation baselines that use cleaner image-text data. When we tune our model on a cleaner subset of image-text data to get \ourclip{}, we see that it sets the absolute state-of-the-art on 4 out of 5 datasets. 
}
\vspace{-15pt}
\label{tab:zsseg}
\end{table}

\begin{SCtable*}[][!t]
\centering
\scalebox{0.74}{%
\begin{tabular}{llclclclclclcl}
\toprule
\textbf{VLM}              & \textbf{Method} & \multicolumn{2}{l}{\textbf{A-847}} & \multicolumn{2}{l}{\textbf{PC-459}} & \multicolumn{2}{l}{\textbf{A-150}} & \multicolumn{2}{l}{\textbf{PC-59}} & \multicolumn{2}{l}{\textbf{VOC-20}} & \multicolumn{2}{l}{\textbf{VOC-21}} \\
\midrule
\texttt{CLIP-B/16} & \texttt{ZegFormer~\citep{ding2022decoupling}}       & 5.6            && 10.4            && 18.0           && 45.5           && 89.5            && 65.5 &            \\
\texttt{CLIP-B/16} & \texttt{ZSseg~\citep{xu2022simple}}           & 7.0            && -             && 20.5           && 47.7           && 88.4            && -  &              \\
\texttt{CLIP-B/16} & \texttt{OVSeg~\citep{liang2023open}}            & 7.1            && 11.0            && 24.8           && 53.3           && 92.6            && -  &              \\
\rowcolor[HTML]{EFEFEF} 
\texttt{CLIP-B/16} & \texttt{CAT-Seg~\citep{cho2023cat}}          & 8.4            && 16.6            && 27.2           && 57.5          && 93.7           && 78.3  &           \\
\rowcolor[HTML]{EFEFEF} 
\textbf{SILC-C-B/16} & \texttt{CAT-Seg~\citep{cho2023cat}}          & 13.4 & \textcolor{ForestGreen}{(+5.0)}             & 22.0 & \textcolor{ForestGreen}{(+5.4)}           & 36.6 & \textcolor{ForestGreen}{(+9.4)}          & 61.2 & \textcolor{ForestGreen}{(+3.7)}          & 95.9 & \textcolor{ForestGreen}{(+2.2)}            & 80.4 & \textcolor{ForestGreen}{(+2.1)}            \\
\rowcolor[HTML]{EFEFEF} 
\textbf{SILC-S-B/16} & \texttt{CAT-Seg~\citep{cho2023cat}}          & 13.5 & \textcolor{ForestGreen}{(+5.1)}             & 21.9 & \textcolor{ForestGreen}{(+5.3)}           & 37.0 & \textcolor{ForestGreen}{(+9.8)}          & 61.2 & \textcolor{ForestGreen}{(+3.7)}          & 96.1 & \textcolor{ForestGreen}{(+2.4)}            & 80.9 & \textcolor{ForestGreen}{(+2.6)}            \\
\midrule
\texttt{CLIP-L/14} & \texttt{ZSseg~\citep{xu2022simple}}           & 7.1            && 10.2            && 21.7           && 52.2           && 92.3            && - &\\
\texttt{CLIP-L/14} & \texttt{OVSeg~\citep{liang2023open}}            & 9.0            && 12.4            && 29.6           && 55.7           && 94.5            && -  &              \\
\rowcolor[HTML]{EFEFEF} 
\texttt{CLIP-L/14} & \texttt{CAT-Seg~\citep{cho2023cat}}          & 10.8           && 20.4            && 31.5           && 62.0           && 96.6            && 81.8 &            \\
\rowcolor[HTML]{EFEFEF} 
\textbf{SILC-C-L/16} & \texttt{CAT-Seg~\citep{cho2023cat}}          & 15.0 & \textcolor{ForestGreen}{(+4.2)}              & 25.8 & \textcolor{ForestGreen}{(+5.4)}           & 37.7 & \textcolor{ForestGreen}{(+6.2)}           & 63.5 & \textcolor{ForestGreen}{(+1.5)}           & 97.6 & \textcolor{ForestGreen}{(+1.0)}            & 82.5 & \textcolor{ForestGreen}{(+0.7)}        \\
\midrule
\texttt{CLIP-G/14} & \texttt{CAT-Seg~\citep{cho2023cat}} & 13.3	&& 21.4 && 36.2 && 61.5 && 97.1 && 81.4 & \\
\bottomrule
\end{tabular}}
\caption{\textbf{Comparing Open Vocabulary Semantic Segmentation performance}, we observe that \ours{} models improve over CLIP by significant margins on all unseen test sets. \ours{} particularly improves the performance for challenging test sets with large vocabularies. SILC-L/16 even outperforms the much larger CLIP-G/14. All models are trained on COCO-Stuff.}
\label{tab:ovseg}
\end{SCtable*}

\subsection{Zero-Shot Semantic Segmentation.}
Zero-shot semantic segmentation aims to measure the grounding performance of a VLM usually from its patch embeddings. 
MaskCLIP~\citep{maskclip} and CLIPpy~\cite{clippy} found that this grounding naturally emerges as a consequence of image-text contrastive training.
We use a Vision Transformer with a MAP pooling head~\citep{scalingvit}. We observe that grounding for our model emerges in the values of the MAP head instead of the last encoder block. 
For a given set of possible classes in a segmentation dataset, we obtain the corresponding text embeddings by querying our text encoder with a standard prompt.
We compute the cosine similarity between the image patch embeddings and the text features of each class name to generate a segmentation map in zero-shot. 
We report the mean-IOU~(mIOU) performance of our model in Table~\ref{tab:zsseg} and compare with baselines at ViT/B16 similar to previous works. We follow the evaluation protocol of TCL~\citep{tcl} without the background class. However, we do not use any post-refinement e.g.~PAMR as we argue that the raw segmentation of a VLM is the true depiction of its zero-shot performance.

\new{\myparagraph{Comparing against CLIP and SigLIP}, we see that both \ourclipp{} and \oursigp{} show significantly superior zero-shot semantic segmentation performance. In fact, both variants achieve multiple mIOU points improvements over all 5 datasets. This validates our hypothesis that the combination of image-text contrastive learning and local-to-global correspondence learning allows the model to develop better understanding of local semantics of the image grounded in language. From  Table~\ref{tab:zsseg}, we observe that the CLIP objective in general results in superior zero-shot segmentation than the SigLIP objective. This is also apparent as we compare \ourclipp{} with \oursigp{}. Moreover, we observe that finetuning on a cleaner subset further improves the zero-shot segmentation performance of both \ourclip{} and \oursig{}. We observe that the CLIP variant \ourclip{} also outperforms \oursig{} here. We show the improvements of \ours{} on CLIP~(WebLI) qualitatively in Figure~\ref{fig:qualitatives_zsseg}. We can observe that \ours{} is better at segmenting and labeling semantic classes in images. We would like to emphasize that \ours{} achieves this without any segmentation ground truth.

\myparagraph{Comparing with current SOTA}, we observe that \ourclip{} consistently beats all specialized zero-shot segmentation baselines on 4/5 datasets to set a new state of the art~(SOTA). Compared to the previous state of the art TCL, \ours{} achieves a remarkable 4.3 mIOU points improvement on A-150, 2.9 points improvement on PC-59, and 4.9 points improvement on CityScapes. Similar improvements are noted on VOC-20 and COCO-Stuff, however Group-VIT maintains the best result on VOC-20. 
These methods use relatively cleaner image captioning datasets for their segmentation specific training objectives. We noticed that the improvements in zero-shot segmentation are achievable by just finetuning on a cleaner subset of data. We did not observe superior performance by learning an expensive patch-wise attention as proposed by PACL~\cite{pacl}. We show in the supplementary that \ours{} models also outperform PACL trained on WebLI.
Methods like TCL and ReCo, which are designed to improve the zero-shot segmentation performance of a frozen VLM, can in theory further improve the performance of our model. However, since we aim to improve vision-language pretraining over all tasks, this is out of the scope of this work.
}

\input{qualitatives/ovseg}

\setlength{\tabcolsep}{4pt}
\renewcommand{\arraystretch}{1.2} 
\begin{table}[t]
\centering
\small
\setlength{\aboverulesep}{0pt}\setlength{\belowrulesep}{0pt}
 \resizebox{0.9\linewidth}{!}
 {%
   \begin{tabular}{ l l  c c c}
  	\toprule
   
  \textbf{Initialization} & \textbf{Training data}	&  \multicolumn{1}{c}{\textbf{COCO}} & \multicolumn{2}{c}{\textbf{LVIS}}\\
  \cmidrule(lr){3-3}  \cmidrule(lr){4-5}  
  & \textbf{}& \textbf{AP} & \textbf{AP\textsubscript{all}} & \textbf{AP\textsubscript{rare}}  \\
  	\midrule
  	\texttt{CLIP~(WebLI)} & WebLI N-grams & 40.4 & 31.9 & 29.2  \\
        \textbf{\ourclipp{}(Ours)} & WebLI N-grams & \textbf{41.8} & \textbf{33.3} & \textbf{30.4}\\
        \midrule
        \texttt{SigLIP} &  WebLI N-grams & 40.9 & 32.8 & 30.4\\
        \textbf{\oursigp{}(Ours)} &  WebLI N-grams & \textbf{42.7} & \textbf{34.2} & \textbf{32.4}\\
  	\bottomrule
  \end{tabular}
}
\caption{\textbf{Training OWLv2 for Object Detection} \new{with \ours{} models offers consistent improvement over CLIP and SigLIP for open vocabulary object detection. These models are trained with pseudo labels from WebLI N-grams~\citep{owlvit} and evaluated zero-shot on COCO and LVIS.}
}
\label{tab:owlvit}
\end{table}

\setlength{\tabcolsep}{4pt}
\renewcommand{\arraystretch}{1.2} 
\begin{table}[t]
\centering
\small
\setlength{\aboverulesep}{0pt}\setlength{\belowrulesep}{0pt}
 \resizebox{\linewidth}{!}
 {%
   \begin{tabular}{ l c c c c c c }
  	\toprule
   
  \textbf{Model}	&  \multicolumn{2}{c}{\textbf{Classification}} & \multicolumn{1}{c}{\textbf{Captioning}} & \multicolumn{2}{c}{\textbf{Question Ans.}}\\
  \cmidrule(lr){2-3}  \cmidrule(lr){4-4}  \cmidrule(lr){5-6} 
  & \textbf{ImageNet} & \textbf{SUN397} & \textbf{COCO} 
  & \textbf{GQA} & \textbf{VQAv2}  \\
  	\midrule
  	\texttt{CLIP~(WebLI)} & 82.3 & 82.4 & 118.1 & 
        52.5 & 63.5 \\
        \textbf{\ourclipp{}(Ours)} & \textbf{83.8} & \textbf{83.4} & \textbf{120.8} & 
        \textbf{53.1} & \textbf{64.6}\\
        \midrule
        \texttt{SigLIP} & 82.5 & 82.2 & 117.5 & 
        51.9 & 63.0 \\
        \textbf{\oursigp{}(Ours)} & \textbf{83.7} & \textbf{82.9} &  \textbf{121.2} &  
        \textbf{53.2} & \textbf{64.5} \\
  	\bottomrule
  \end{tabular}
}
\caption{\textbf{Evaluating \ours{} visual representation} \new{with LiT-Decoder in a multi-task setup, we observe consistent improvements on all tasks compared to CLIP and SigLIP. These improvements are especially apparent for tasks that require local understanding of the image i.e. Captioning and Question Answering.}
}
\label{tab:litdecoder}
\end{table}

\setlength{\tabcolsep}{4pt}
\renewcommand{\arraystretch}{1.2} 
\begin{table*}[t]
\centering
\small
\setlength{\aboverulesep}{0pt}\setlength{\belowrulesep}{0pt}
 \resizebox{\linewidth}{!}
 {%
   \begin{tabular}{ l c c c c c c  c c c c c c c}
  	\toprule
   
  \textbf{Model}	& \textbf{ImageNet 0 shot} & \multicolumn{3}{c}{\textbf{ImageNet Few shot}} & \multicolumn{2}{c}{\textbf{COCO Retrieval}} & \multicolumn{3}{c}{\textbf{ZS Segmentation}} & \multicolumn{3}{c}{\textbf{Open Vocab Seg}}\\
  \cmidrule(lr){2-2}  \cmidrule(lr){3-5}  \cmidrule(lr){6-7}  \cmidrule(lr){8-10} \cmidrule(lr){11-13} 
  & \textbf{T1} & \textbf{1shot} & \textbf{5shot} & \textbf{10shot} & \textbf{I2T@1} & \textbf{T2I@1} & \textbf{A-150} & \textbf{Stuff} &\textbf{PC-59} & \textbf{PC-459} & \textbf{A-150} & \textbf{PC-59} \\
  	\midrule
  	\texttt{CLIP (WebLI)} & 71.7 & 36.4 & 57.7 & 62.5  & 59.1 & 42.9 & 11.8 & 12.9 & 20.1 & 18.6 & 30.5 & 57.7 \\ 
        \texttt{+ additional views} & 73.6 & 38.7 & 60.8 & 65.7 & 60.6 & 43.2 & 11.7 & 13.0 & 20.0 & 19.2 & 32.1 & 57.8\\
        \texttt{+ EMA} & 73.7 & 38.4 & 60.7 & 65.5 & 61.3 & 43.1 & 11.9 & 13.3 & 20.5 & 19.0 & 32.2 & 57.5\\
        \texttt{+ Self Dist} (\textbf{\ourclipp{}}) & \textbf{74.3} & \textbf{39.9} & \textbf{61.2} & \textbf{65.7} & \textbf{62.7} & \textbf{43.9} & \textbf{12.2} & \textbf{15.3} & \textbf{21.1} & \textbf{21.0} & \textbf{33.3} & \textbf{60.7}\\
  	\bottomrule
  \end{tabular}
}
\caption{\textbf{We ablate over each component} of our model to verify our design choices. The addition of image augmentation and EMA to CLIP~(WebLI) improves classification and retrieval metrics while only slightly impact the segmentation. Adding local-to-global consistency by self-distillation, we observe an improvement across the board especially on segmentation metrics.
}
\label{tab:ablation}
\end{table*}

\subsection{Open-Vocabulary Semantic Segmentation.}
Open Vocabulary Semantic Segmentation aims to develop segmentation models that can segment novel classes beyond the training vocabulary.
Most recent methods in this area rely on a pretrained CLIP due to its open-vocabulary capabilities and adapt it for segmentation task. To evaluate the open vocabulary segmentation potential of \ours{}, we take the current state-of-the-art model CAT-Seg~\citep{cho2023cat} and replace the CLIP model used by the authors with \ours{}. The models are trained on COCO-Stuff-164k with 172 classes and tested on unseen datasets with different vocabularies: ADE-20k with 847 or 150 classes (A-847/A-150), Pascal Context (PC-459/PC-59), and Pascal VOC (VOC-20/VOC-21).

From Table~\ref{tab:ovseg}, we observe that \ours{} significantly improves over CLIP~\cite{clip}. 
In fact, SILC-C-B/16 performs on par with the much bigger CLIP-G/14 on the three most challenging test datasets A-847, PC-459 and A-150. Moreover, we observe that while \oursig{} performed slightly worse than \ourclip{} in zero-shot segmentation, it achieves slightly better performance when trained for open vocabulary segmentation. SILC-S-B/16 further improve on the performance of SILC-C-B/16.
The observed improvements of \ourclip{} also transfer to the larger ViT-L variant, where CAT-Seg with SILC-C-L/16 outperforms CAT-Seg with CLIP-L/14 on all datasets by a significant margin. In particular, it achieves more than +4 mIOU improvement on the challenging A-847, PC-459, and A-150. SILC-L/16 even significantly outperforms the much bigger CLIP-G/14 on all tested datasets.
The improvements of \ourclip{} over CLIP are also reflected in the qualitative examples in Fig.~\ref{fig:qualitatives_ovseg}. We observe that \ourclip{} better distinguishes semantically similar classes such as grandstand/building, field/grass, runway/road and grandstand/chair. Further, it improves segmentation in difficult cases and better handles transparent segments as shown in supplementary. \new{Results on additional WebLI models are also provided in supplementary with similar conclusions.} 

\subsection{\ours{} for Open-Vocabulary Detection.}
\new{We utilize OWLv2~\citep{owlvit} as a framework to test the open vocabulary detection potential of \ours{} models. OWLv2 initializes a detection model with a contrastive image-text model's weights and utilizes pseudo labelled boxes from WebLI~(WebLI N-grams~\citep{owlvit}) to learn an open vocabulary detection model. We utilize the test bench of the authors and retrain OWLv2 initialized from our VLM baselines to report results in Table~\ref{tab:owlvit}. We evaluate these models zero-shot on COCO and LVIS without doing any finetuning on respective dataset to test their open vocabulary performance. 
From Table~\ref{tab:owlvit} we observe that \ours{} models also benefit open vocabulary detection thanks to learning better local semantics.
\oursigp{} achieves an improvement of +1.8AP on COCO. The improvements are also consistent on the challenging LVIS benchmark where \oursigp{} achieves an improvement of +1.4AP on all classes and a remarkable +2.0AP on rare classes. We make similar observations as we compare \ourclipp{} with CLIP~(WebLI) where \ourclipp{} offers consistent improvements. This further validates that \ours{} models offer better performance for dense tasks.}

\subsection{Evaluating \ours{} features with LiT-Decoder.}
\new{LiT-Decoder~\citep{litdecoder} proposes to utilise a frozen image encoder and train a single autoregressive decoder to learn a multi-task model for Classification, Captioning and Visual Question Answering. We use LiT-Decoder as a framework to evaluate the quality of visual representations learned by \ours{} models against baseline CLIP~(WebLI) and SigLIP. We use the authors' implementation and only replace the ViT with the respective baselines. We report results in Table~\ref{tab:litdecoder}. 
We observe that \ours{} models offer consistent improvements in this multi-task setup as well.
Compared to SigLIP, \oursigp{} improves classification by +1.2 points on ImageNet and +1.0 point on SUN397. The improvements are even more profound on captioning~(+3.6 CIDEr score on COCO) and Visual Question Answering~(+1.3 on GQA and +1.5 on VQAv2). These tasks greatly benefit from \ours{} features' ability to better encode local semantics. Similar improvements are noted for CLIP based baselines. This further validates that \ours{} features can simultaneously benefit multiple computer vision problems.}

\subsection{Ablation on Model Components.}
\label{sec:ablation}
We ablate on the various design choices of our model and their impact on various tasks. We train all models for 5 Billion example-seen and report the performance in Table~\ref{tab:ablation}. Since our method processes additional image augmentations in the contrastive loss, we first test if our improvements are a consequence of processing more augmentations. We observe that the introduction of additional image augmentations~(second row) improve the classification and retrieval metrics but their impact on zero-shot segmentation and open vocabulary segmentation is not as significant. When we add an EMA over this model's weights similar to our model~(third row), we notice a slight improvement as seen in previous SSL literature. Finally when we add the self-distillation from local crops, we see an improvement across the board on all tasks. 
In particular, we observe the strongest improvement on segmentation tasks highlighting our proposal's impact on them.

\section{Conclusion.}
\new{We propose to integrate local-to-global correspondence learning by self-distillation as a complementary objective to the popular VLM contrastive objective of CLIP~\citep{clip} and SigLIP~\citep{zhai2023sigmoid}. 
We show that the introduction of this results in remarkable performance improvements on several computer vision tasks.
We see a consistent performance improvement on zero-shot classification, few-shot classification, and retrieval. We further test our VLM on zero-shot segmentation and show that our training framework results in significant improvements without using any dense ground truth. Finally we show that \ours{} models as pretrained backbones significantly improve a model's performance on open vocabulary segmentation, open vocabulary detection, captioning and VQA. \ours{} models set a new state of the art in Vision-Language Foundational Models.}

{
    \small
    \bibliographystyle{ieeenat_fullname}
    \bibliography{main}
}

\end{document}

%% file: preamble.tex
\usepackage[dvipsnames]{xcolor}

\newcommand{\new}[1]{{#1}}

%% file: qualitatives/zsseg.tex
\begin{figure*}[t]
\centering
\small
\setlength{\tabcolsep}{1pt}
\begin{tabular}{ccc}
 &
\begin{tabularx}{.47\linewidth}{*{4}{Y}}
Image & CLIP & SILC-C & G. Truth \\
\end{tabularx} &
\begin{tabularx}{.47\linewidth}{*{4}{Y}}
Image & CLIP & SILC-C & G. Truth \\
\end{tabularx} \\

\rotatebox[origin=l]{90}{\hspace{4mm}A-150} &
\includegraphics[width=.47\linewidth]{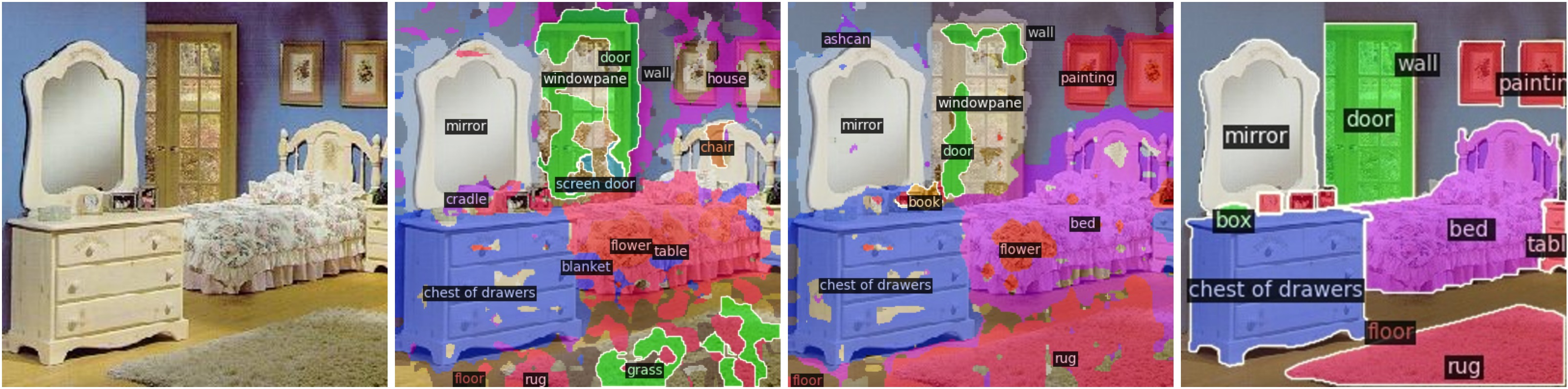} &
\includegraphics[width=.47\linewidth]{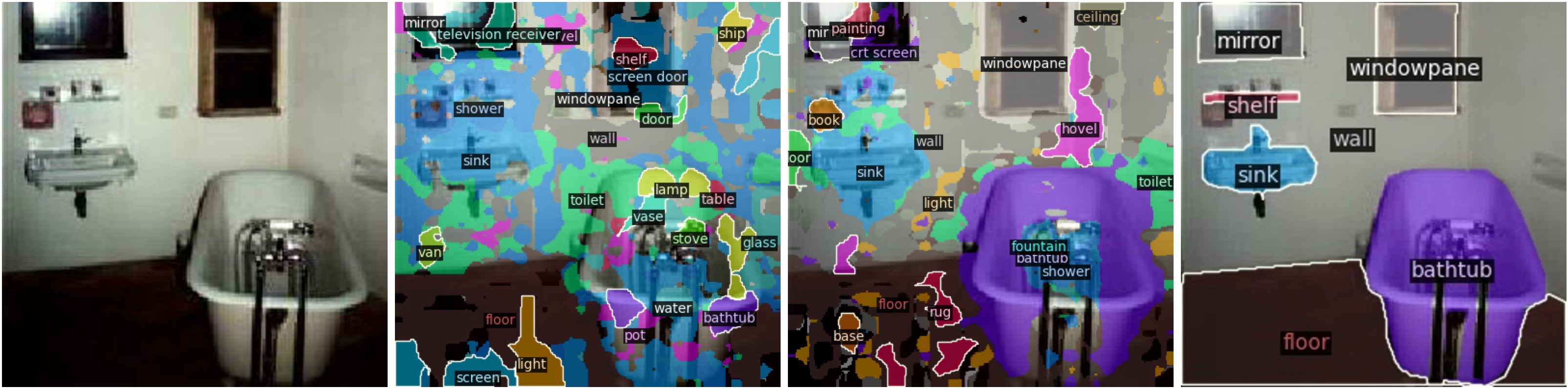} \\[-3pt]

\rotatebox[origin=l]{90}{\hspace{5mm}PC-59} &
\includegraphics[width=.47\linewidth]{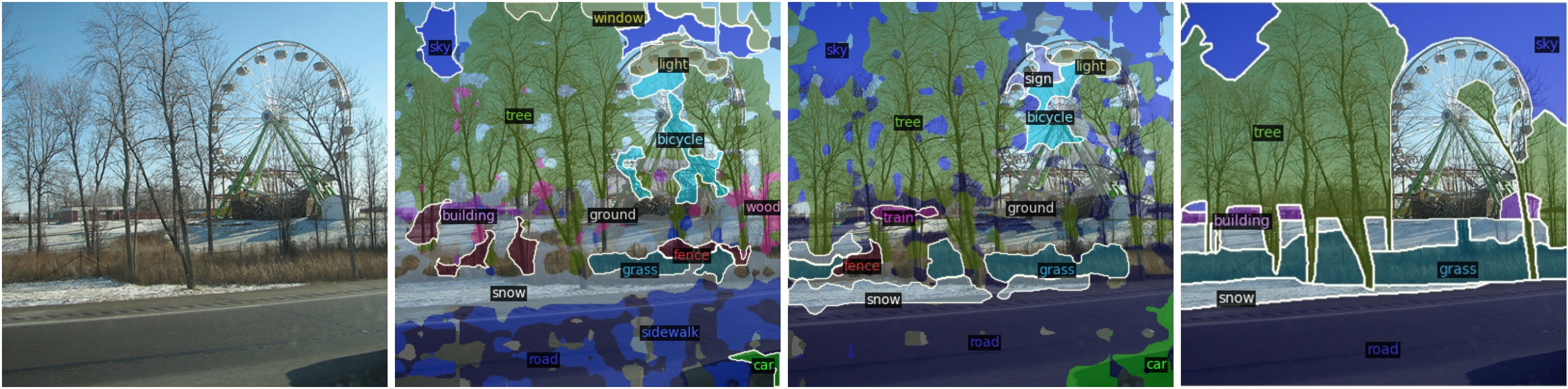} &
\includegraphics[width=.47\linewidth]{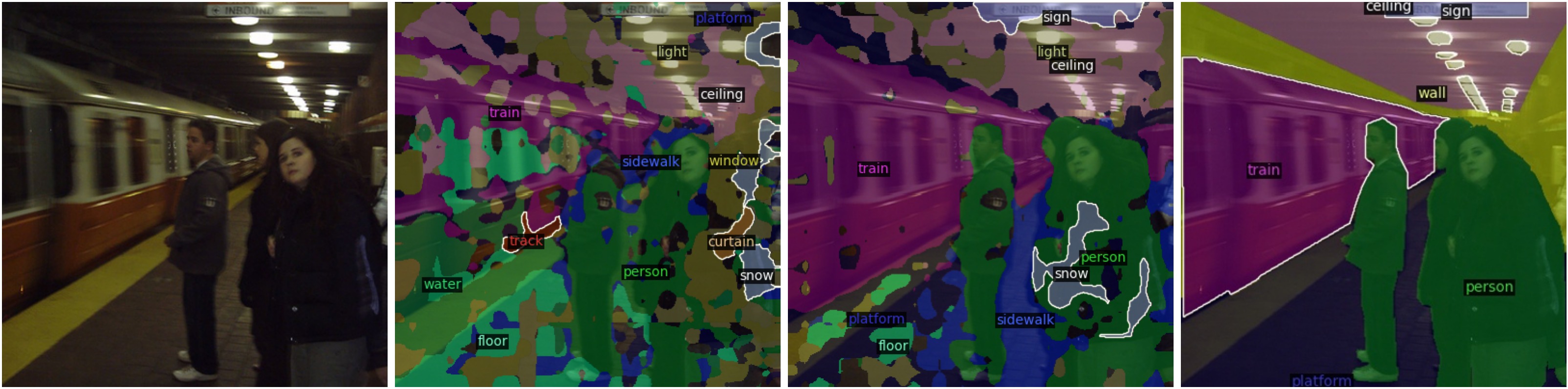} \\[-3pt]
\end{tabular}

\caption{
\textbf{Qualitative results on zero-shot segmentation} show that \ourclip{} achieves significant improvements over CLIP~(WebLI). \ourclip{} produces less noisy segmentation and better distinguishes semantic classes. This semantic segmentation emerges without any segmentation supervision.
}
\vspace{-10pt}
\label{fig:qualitatives_zsseg}
\end{figure*}

%% file: qualitatives/ovseg.tex
\begin{figure*}[t]
\centering
\footnotesize
\setlength{\tabcolsep}{1pt}
\begin{tabular}{ccc}
&
\begin{tabularx}{.47\linewidth}{*{4}{Y}}
Image & CLIP & SILC-C & G.Truth \\
\end{tabularx} &
\begin{tabularx}{.47\linewidth}{*{4}{Y}}
Image & CLIP & SILC-C & G.Truth \\
\end{tabularx} \\
\rotatebox[origin=l]{90}{\hspace{4mm}A-150} &
\includegraphics[width=.47\linewidth]{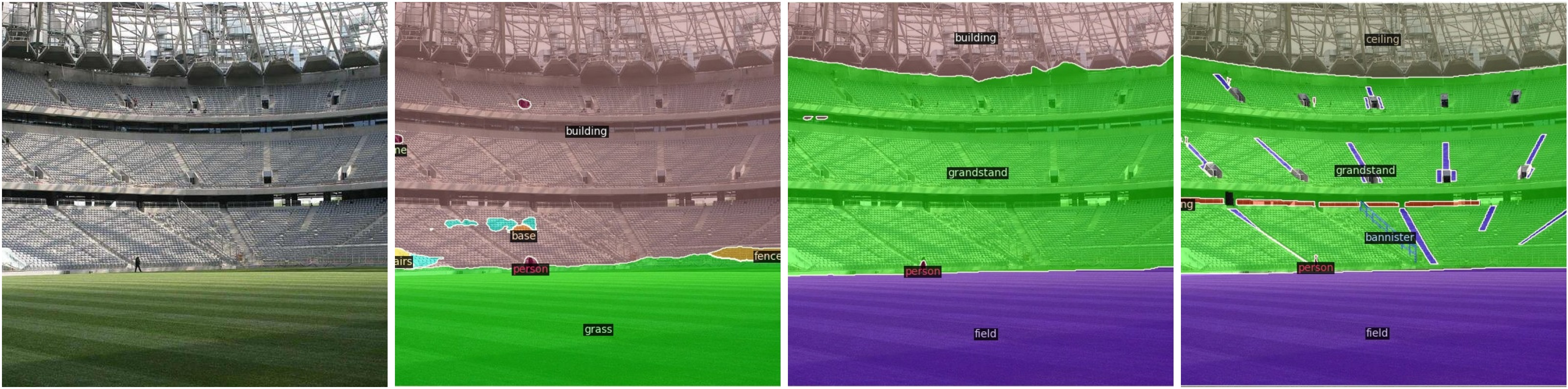} &
\includegraphics[width=.47\linewidth]{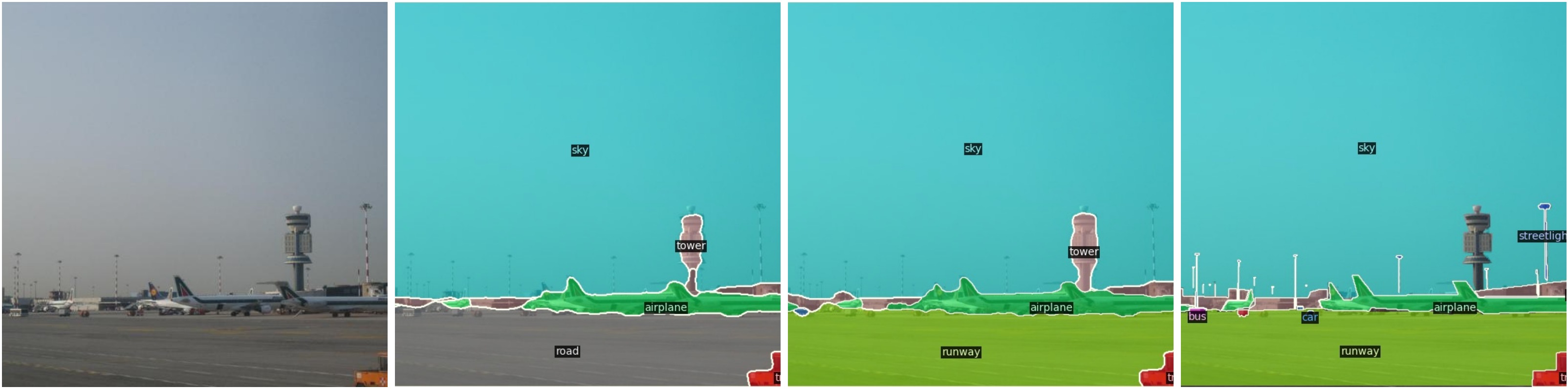} \\[-3pt]

\rotatebox[origin=l]{90}{\hspace{4mm}PC-459} &
\includegraphics[width=.47\linewidth]{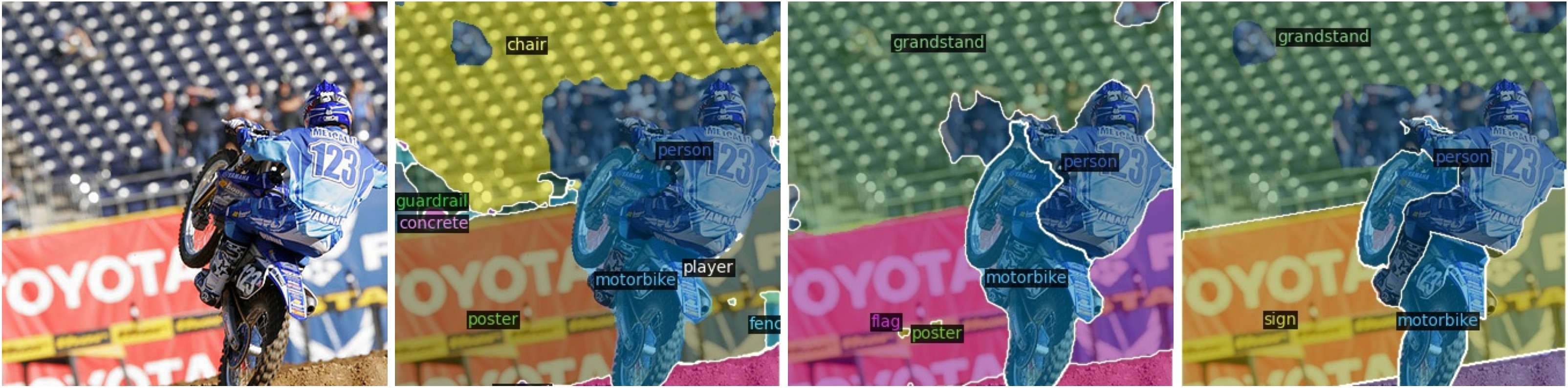} &
\includegraphics[width=.47\linewidth]{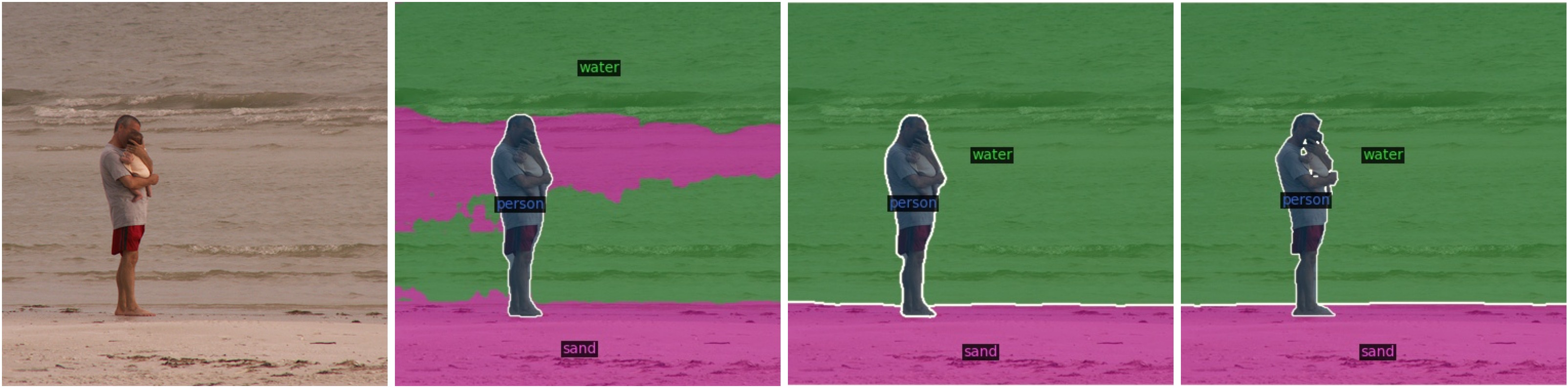} \\[-3pt]
\end{tabular}

\caption{\textbf{Comparing qualitative examples for open vocabulary segmentation}, we observe that \ours w/ CAT-Seg better distinguishes semantically similar classes such as field/grass, runway/road, grandstand/chair and sand/water than CLIP.}
\vspace{-5pt}
\label{fig:qualitatives_ovseg}
\end{figure*}